\documentclass{article}

\usepackage[utf8]{inputenc}
\usepackage[english]{babel}
\usepackage[dvipsnames]{xcolor}
\usepackage[final]{nips_2018}
\PassOptionsToPackage{numbers, compress}{nonatbib}
\usepackage{hyperref}
\usepackage{amssymb, amsmath}
\usepackage{url}
\usepackage{graphicx}
\usepackage{subcaption}
\usepackage{wrapfig, float}
\usepackage{tabularx,booktabs}
\newcolumntype{C}{>{\centering\arraybackslash}X} 
\usepackage{xcolor}
\usepackage{color, colortbl}
\usepackage{tabularx} 
\usepackage[dvipsnames]{xcolor}
\usepackage[backend=biber,style=ieee, maxnames=4]{biblatex}
\usepackage{csquotes}

\usepackage{siunitx}
\renewrobustcmd{\bfseries}{\fontseries{b}\selectfont}
\renewrobustcmd{\boldmath}{}
\newrobustcmd{\B}{\bfseries}
\newcommand{\Methodnametitle}{Making distance matter in high dimensional biological analysis}
\newcommand{\Methodnamelong}{METric learning for Confounder Control}
\newcommand{\Methodname}{METCC}

\newcommand{\SampleTotal}{817}
\newcommand{\NumberGenes}{24152}

\definecolor{rowgold}{HTML}{F6ED8D}
\definecolor{rowred}{HTML}{FF7777}
\definecolor{rowgreen}{HTML}{B7DC81}

\title{\Methodname: \Methodnamelong\\
\Methodnametitle}

\author{
Kabir Manghnani\\
\texttt{kabir.manghnani@freenome.com}\\
\texttt{ kabirm2@illinois.edu}\\
\And
Adam Drake\\
\texttt{adam.drake@freenome.com}
\\
\And
Nathan Wan\\
\texttt{nwan@freenome.com}
\And
Imran S. Haque\\
\texttt{ihaque@cs.stanford.edu}\
}

\begin{document}

\maketitle

\begin{abstract}

High-dimensional data acquired from biological experiments such as next-generation sequencing are subject to a number of confounding effects.  These effects include both technical effects, such as variation across batches from instrument noise or sample processing ("batch effects"), or institution-specific differences in sample acquisition and physical handling ("institutional variability"), as well as biological effects arising from true but irrelevant differences in the biology of each sample, such as age biases in diseases. Prior work has used linear methods to adjust for such batch effects. Here, we apply contrastive metric learning by a non-linear triplet network to optimize the ability to distinguish biologically distinct sample classes in the presence of irrelevant technical and biological variation. Using whole-genome cell-free DNA data from \SampleTotal\ patients, we demonstrate that our approach, \Methodnamelong \ (\Methodname), is able to match or exceed the classification performance achieved using a best-in-class linear method (HCP) or no normalization. Critically, results from METCC appear less confounded by irrelevant technical variables like institution and batch than those from other methods even without access to high quality metadata information required by many existing techniques; offering hope for improved generalization.
\end{abstract}

\section{Introduction}
The acquisition of biological data, such as DNA sequencing, requires complex and inter-dependent
procedures that are sensitive to variables that are not of interest to the data's user \cite{qiagen}. 
Numerous factors, including but not limited to sample processing, storage time, and temperature, all affect the simultaneous measurement of thousands to millions of variables in DNA sequencing.
These confounders pose a challenge when developing solutions to pattern
recognition problems using biological data because they can obscure the biological signal of
interest.

Accounting for and normalizing technical variables from data has a rich history dating back at least
100 years to RA Fisher's work on ANCOVA and random effects \cite{ancova, randomeffects}. 
Many normalization methods are derived from this initial formulation that models the response variable as a function of both global and group level
parameters. These models specify a mean effect parameter to be estimated per group that accounted for
unwanted group-wise variation of the mean. More flexible models have been developed: mixed effects
models, for example, fit a set of global and 
group indexed parameters \cite{combat, hidden, HCP}. In particular, HCP
(Hidden Covariates with Prior) models the normalized data, $\hat{Y}$, as $N(\hat{Y}|XW+FB, I)$ where $F$ 
is a matrix of known covariates, $X$ is a matrix of unknown covariates, and $W$, $B$, and $X$ are estimated
from the data \cite{HCP}.
Depending on the hyperparameters, this model subsumes model specification similar to two popular normalization models: ComBat \cite{combat} and PANAMA \cite{panama}.  Separately, there are efforts in to estimate data representations
that are invariant with respects to technical effects. For example, Shaham et al. normalizes
confounder effects in RNA sequencing using a variational adverserial approach
\cite{blackboxbatch}. The application of their method is unfortunately limited to pairs of
batches.

Unlike previous models, we approach the covariate normalization problem
as a reduced representation learning problem using methods of blackbox
metric learning such as siamese and triplet networks \cite{Hadsell, triplet}. By enforcing a learned metric
with a loss function aiming to preserve biological information we learn a representation that
contains less of the variation due to confounding effects. Metric learning methods are advantageous because
the loss function requires only the variable of interest as opposed to mixed effects models which also require
annotation of the technical effects to normalize. Additionally our method provides the ability to model non-linear effects unlike that of HCP.

We analyze 1) the extent to which data normalized with HCP and \Methodname\ retains information about the
unwanted technical effects and 2) the performance of supervised models trained on normalized data.


\section{Methods}
\subsection{\Methodname\ learning} \label{METCC_learning}

Let $X$ be an $n \times p$ matrix of observed biological data with $n$ samples and $p$ measurements. 
Let $y$ be a biological variable of interest such as phenotype label or disease status.
We seek to learn a distance function $D_w$ parameterized by the map $g: \mathbb{R}^p \rightarrow \mathbb{R}^k$ 
where the distance between 2 samples $x_i$ and $x_j$ is determined by $D_w = ||g(x_i)-g(x_j)||_2$.
The objective is to transform the data with $g$ 
so that the variability measured between sample $x_i$ and $x_j$ is low when 
$y_i == y_j$ and high when $y_i \neq y_j$. This can be
optimized using a contrastive, or siamese, loss function proposed by Hadsell et al. \cite{Hadsell},

\begin{equation}
    \label{siamese}
    argmin_w{(1-L) D_w(x_i, x_j)^2 + (L)max(0, m - D_w(x_i, x_j))^2 }
\end{equation}

where $L=0$ if $y_i \neq y_j$, $L=1$ otherwise, and $m$ is a fixed margin. Equation
(\ref{siamese}) can be described as follows, the first term of the loss function encourages samples from the same class but different institutes to be close in the latent space; while the second penalizes samples from the same institute that remain close in the latent space while belonging to different disease classes.

This has been extended and shown to produce better representations when both a positive
and negative class sample are used for each anchor sample.
Hoffer and Ailon proposed a triplet loss that
extends equation (\ref{siamese}) by running a triplet of samples $(x^-, x, x^+)$ through $D_w$
where $x$ is
an anchor sample, $x^-$ is from a different and $x^+$ is from the same class \cite{triplet}.
The distances obtained are $d_+=D_w(x, x^+)$ and $d_-=D_w(x, x^-)$ and the loss function minimized is,
\begin{equation}
    \label{triplet}
    argmin_w ||d_+, d_--1||^2_{2}
\end{equation}
This loss function poses the metric learning task as classifying $x^+$ and $x^-$ as either being a member
of the $x$ class or not.

Hoffer and Ailon suggest a distance function defined with a softmax on the outputs $D_w$ \cite{triplet}; we instead use the formulation of Balnter et al. that forgoes the softmax \cite{torchtriplet}.

In practice, the biological data coming off of a sequencer can have tens of thousands of measured variables.
In order to obtain a stable solution to equations (\ref{siamese}) or (\ref{triplet}) with gradient descent we found that
unsupervised dimensionality reduction techniques like principal components analysis (PCA) need to be applied
to $X$ prior to metric learning likely due to our relatively limited sample size. We trained $D_w$ where $X$ was a matrix of cfDNA data and $y$ was a label to predict.

\begin{figure}
    \centering
    \includegraphics[scale=0.2]{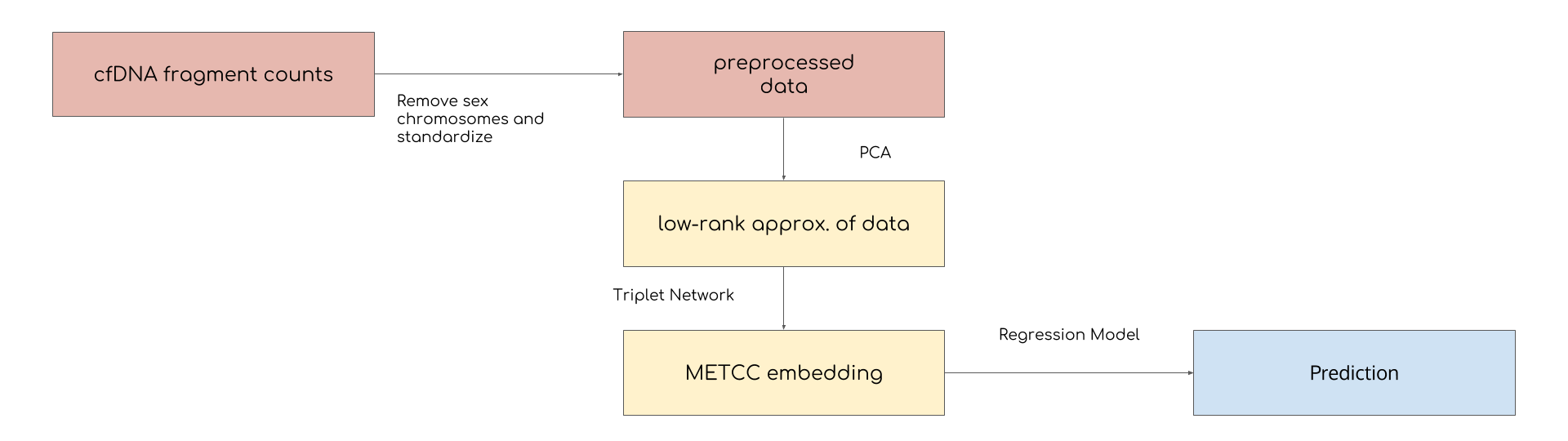}
    \caption{\Methodname\ data workflow }
    \label{fig:model_workflow}
\end{figure}

\subsection{Classifications} \label{clf}
Once $D_w$ is trained we take the output of $g(.)$ as the METCC embedding for each sample 
(see workflow in Figure \ref{fig:model_workflow}).  In order to assess the quality of these \Methodname\ embeddings, separate embeddings are also constructed from baseline methods of PCA and HCP from the same samples, and all three embeddings are subsequently used to train both a k-nearest neighbor (KNN) and a logistic regression (LR) model to predict disease with k-fold cross validation $(k=4)$ \cite{scikit-learn}.  In order to assess the confounder signals in the embeddings, regression models are also trained with confounder values as the target label.

\section{Experiments and results}
Our dataset is a collection of \SampleTotal\ cell-free DNA (cfDNA) samples and their associated metadata \cite{freenome}.
The raw data $X$ consists of the number of DNA fragments overlapping each gene annotated in the CHESS gene
set consisting of \NumberGenes\ gene features for each sample \cite{chess}. All samples were sequenced to approximately 10x depth (i.e., each base in the genome appeared, on average, in approximately 10 independent fragments). Each sample is labeled as coming from an individual with colorectal cancer or from a healthy individual. Positive labels were confirmed by colonoscopy and expert analysis of pathology or histology
reports.

\subsection{Generating Embeddings} \label{gen_embeddings}

In order to assess the quality of \Methodname\ embeddings, we prepared two other embeddings to compare against:
\begin{equation}
\label{embedding}
\begin{split}
    X_{PCA} &= PCA(x) \\
    X_{HCP} &= PCA(HCP(X)) \\
    X_{\Methodname} &= \Methodname(PCA(X))
\end{split}
\end{equation}

Before $X$ is transformed a basic pre-processing step is applied to the raw data that removes sex chromosomes and standardizes each sample. We apply PCA after HCP in order to have comparable
dimensions for the data used when training a classifier.
$X_{PCA}$ is a baseline comparator with no additional normalization, just dimensionality reduction with PCA.
$X_{HCP}$ is embedding normalized with HCP using only institution, batch, and age labels. The institution reflects the provenance of
the patient's sample. The batch entails the sequencing chemistry preparation and process
and thus reflects both short and long term effects in the data. We grouped age into the
bins: [0-50, 50-55, 55-60, 60-75, 75-80, 80-85, 85+]. $X_{\Methodname}$ is the METCC embedding that is the result of the method described in \ref{METCC_learning}. \Methodname\ has no knowledge of any confounders, only the disease label. Hyperparameters for HCP and \Methodname\  were selected by choosing the best performing logistic regression model across a random search of the normalization methods' hyperparameters and regularization of the logistic regression model (see section \ref{hyperparameter_selection}).  Regularization for all logistic regression models including confounder tasks were performed independently.

Manifolds of these embeddings are plotted by various targets in section \ref{manifolds}.

\subsection{Classification and embedding analysis}

For each set of embeddings in section \ref{gen_embeddings},  we first classify by disease label, the target of biological interest.  The mean of the Area Under the Receiver Operating Characteristics (AUROC) for each test fold is used as an estimate for performance.  As shown previously \cite{freenome}, logistic regression can perform very well on this task, even with just PCA; METCC performs comparably when using logistic regression, but definitely stands out when using a non-parametric, KNN model ($k=21$, selected by minimizing the difference in train/test accuracy).

\begin{figure}
    \centering
    \includegraphics[scale=0.3]{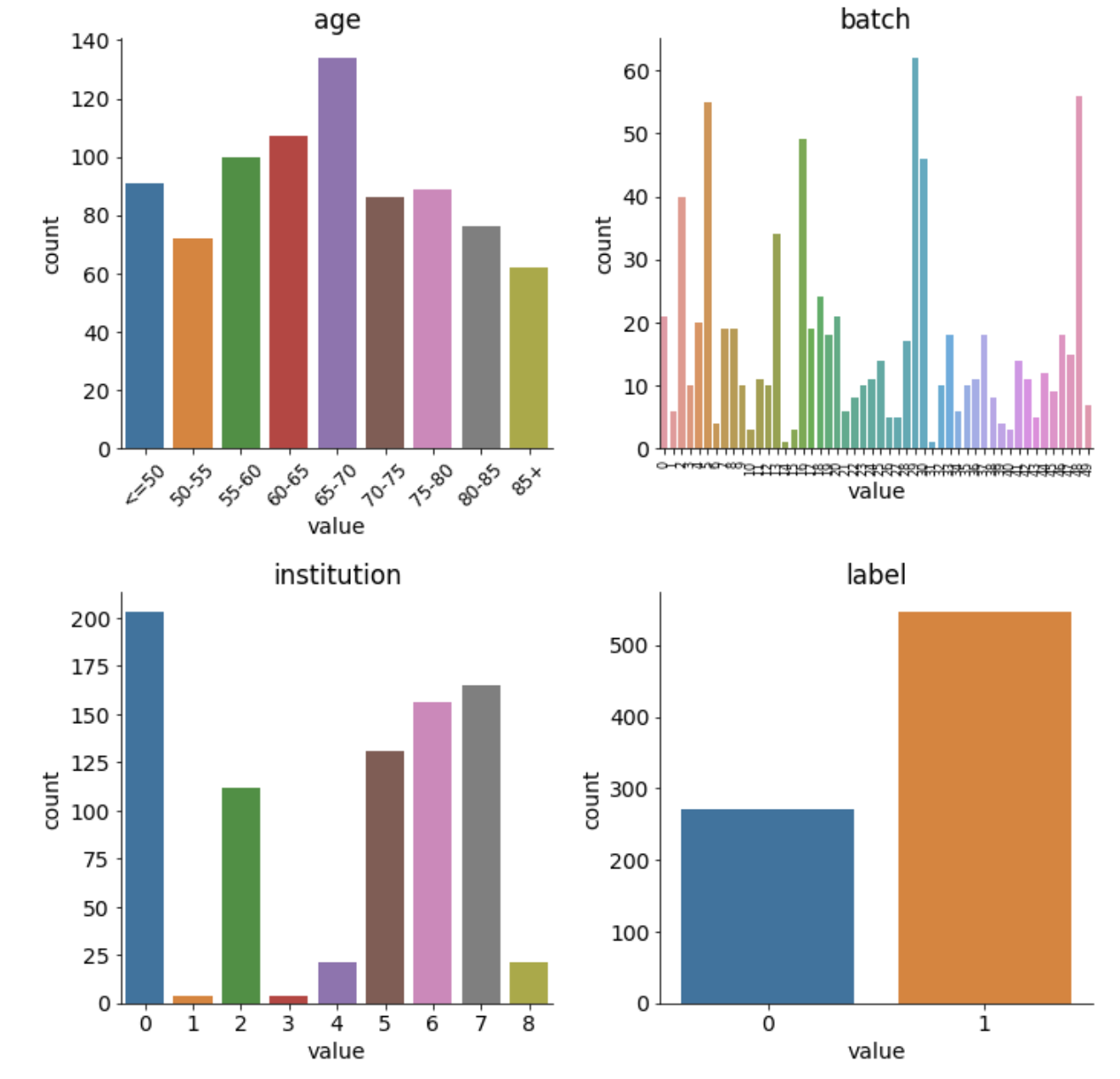}
    \caption{Confounder label distribution}
    \label{fig:confounder_distribution}
\end{figure}

In addition to disease label classification, we also assess the signal of confounders described in section \ref{gen_embeddings} (plotted in Figure \ref{fig:confounder_distribution})
by training classifiers using the confounders as the target label value.  Since this is no longer a binary classification problem, performance is measured by mean accuracy.  Although sometimes subtle, embeddings produced by \Methodname\ are significantly more difficult to predict confounders on when the confounder is technical, that is either institution or batch; the removal of unwanted signal is evident across both regression methods.  The difference between the normalization methods is not dramatic for age, but from the confounder distribution
it is clear that an increase in age is correlated with increased likelihood of disease.

\begin{table*}
  \centering
  \begin{tabular}{l|rr|rr}
    \toprule
    \textbf{Normalization} & \textbf{Train AUC (KNN)} & \textbf{Test AUC (KNN)} 
    & \textbf{Train AUC (LR)} & \textbf{Test AUC (LR)} 
    \\
    \midrule
    PCA-only & \num{0.865} $\pm$ \num{0.00802411} & \num{0.788622} $\pm$ \num{0.0359613} & \num{0.973445} $\pm$ \num{0.00238017} & \num{0.924844} $\pm$ \num{0.0129481} \\
    HCP  & \num{0.900041} $\pm$ \num{0.00792789}  & \num{0.824976} $\pm$ \num{0.0395908} & \num{0.98585} $\pm$ \num{0.00125681} & \num{0.914198} $\pm$ \num{0.0124774} \\
    \Methodname & \num{0.999265} $\pm$ \num{0.000321651} & \B \num{0.874324} $\pm$ \num{0.0230618} & \num{0.997703} $\pm$ \num{0.000665161} & \num{0.922061} $\pm$ \num{0.00963767}\\
    \bottomrule
    \end{tabular}
      \caption{Mean k-fold AUROC of disease label classification for different normalization techniques using
      K-nearest neighbors (KNN) and logistic regression (LR) across folds.}
  \label{table:classificantion_stats}
\end{table*}

\begin{table*}
  \centering
  \begin{tabular}{l|rr|rr}
    \toprule
    \textbf{Normalization}  & \textbf{Train ACC (KNN)} & \textbf{Test ACC (KNN)} &
    \textbf{Train ACC (LR)} & \textbf{Test ACC (LR)}\\
   \midrule
    
    PCA-only (inst)  & 
        \num{0.529993} $\pm$ \num{0.004171} & 
        \num{0.241083} $\pm$ \num{0.026689} & 
        \num{0.823339} $\pm$ \num{0.006507} & 
        \num{0.304708} $\pm$ \num{0.053285}\\
    HCP (inst)  & 
        \num{0.544266} $\pm$ \num{0.017817} & 
        \num{0.198136} $\pm$ \num{0.050606} & 
        \num{0.901270} $\pm$ \num{0.005377} & 
        \num{0.309592} $\pm$ \num{0.046238}\\
    \Methodname\ (inst)  & 
        \num{0.504287} $\pm$ \num{0.007636} & 
        \num{0.226395} $\pm$ \num{0.072604} & 
        \B \num{0.383916} $\pm$ \num{0.009528} & 
        \B \num{0.221668} $\pm$ \num{0.094189} \\
    
    \midrule
    PCA-only (batch) & 
        \num{0.286001} $\pm$ \num{0.006259} & 
        \num{0.062411} $\pm$ \num{0.029521} & 
        \num{0.643407} $\pm$ \num{0.015667} & 
        \num{0.085642} $\pm$ \num{0.017647}\\
    HCP (batch) & 
        \num{0.194231} $\pm$ \num{0.028331} & 
        \num{0.036717} $\pm$ \num{0.002407} & 
        \num{0.904527} $\pm$ \num{0.005510} & 
        \num{0.093019} $\pm$ \num{0.003315}\\
    \Methodname\ (batch) & 
        \num{0.218266} $\pm$ \num{0.013615} & 
        \num{0.070966} $\pm$ \num{0.018971} & 
       \B \num{0.116684} $\pm$ \num{0.011186} & 
        \B \num{0.057569} $\pm$ \num{0.012421}\\
    
    \midrule
    PCA-only (age) & 
        \num{0.365554} $\pm$ \num{0.010306} & 
        \num{0.121146} $\pm$ \num{0.017561} & 
        \num{0.621374} $\pm$ \num{0.007255} & 
        \num{0.122311} $\pm$ \num{0.038451}\\
    HCP (age) & 
        \num{0.276182} $\pm$ \num{0.029898} & 
        \num{0.145649} $\pm$ \num{0.039486} & 
        \num{0.752342} $\pm$ \num{0.005851} & 
        \num{0.136975} $\pm$ \num{0.040510}\\
    \Methodname\ (age) & 
        \num{0.310890} $\pm$ \num{0.007536} & 
        \num{0.116273} $\pm$ \num{0.008657} & 
        \num{0.233775} $\pm$ \num{0.010438} & 
        \num{0.129784} $\pm$ \num{0.044002}\\

       \bottomrule
    \end{tabular}
      \caption{Mean k-fold accuracy of confounder label classification for institution (inst), batch, and age for different normalization techniques using
      K-nearest neighbors (KNN) and logistic regression (LR) across folds.}
  \label{table:cpt_stats}
\end{table*}

\section{Discussion and conclusion}
We propose the use of a blackbox metric learning method for confounder normalization that accounts
for confounder effects without the necessity of having \textit{a priori} knowledge of these
effects, which can be rare or hard to procure. \Methodname\ can outperform data minimally pre-processed and data normalized with a mixed effects model in a biological prediction task as measured by AUROC 
(Table \ref{table:classificantion_stats}). 
We analyzed the data above in confounder
prediction tasks in order to measure a proxy for preserved covariate signal
in the data (Table \ref{table:cpt_stats}) and observed that the minimally pre-processed and data
normalized with HCP consistently predict at or better than \Methodname. This indicates that \Methodname\ may be abrogating more of the counfounder's effects than compared methods.

Our work demonstrates the feasibility of mitigating non-linear interactions
in data that do not facilitate solving a desired pattern recognition task in high dimensional
biological data. Future studies will compare black box metric learning models
to non-linear mixed effects models as well as assessing more directly the removal of confounder
effect and the highlighting of biological signal in learned transformations.


\subsubsection*{Acknowledgments}

The authors gratefully acknowledge Signe Fransen, Girish Putcha and all colleagues at Freenome for their extensive suggestions, feedback, and editorial support. 

\printbibliography
\clearpage

\section{Supplemental materials}
\label{supplement}
\subsection{Label distribution}
\begin{figure}
    \centering
    \includegraphics[scale=0.3]{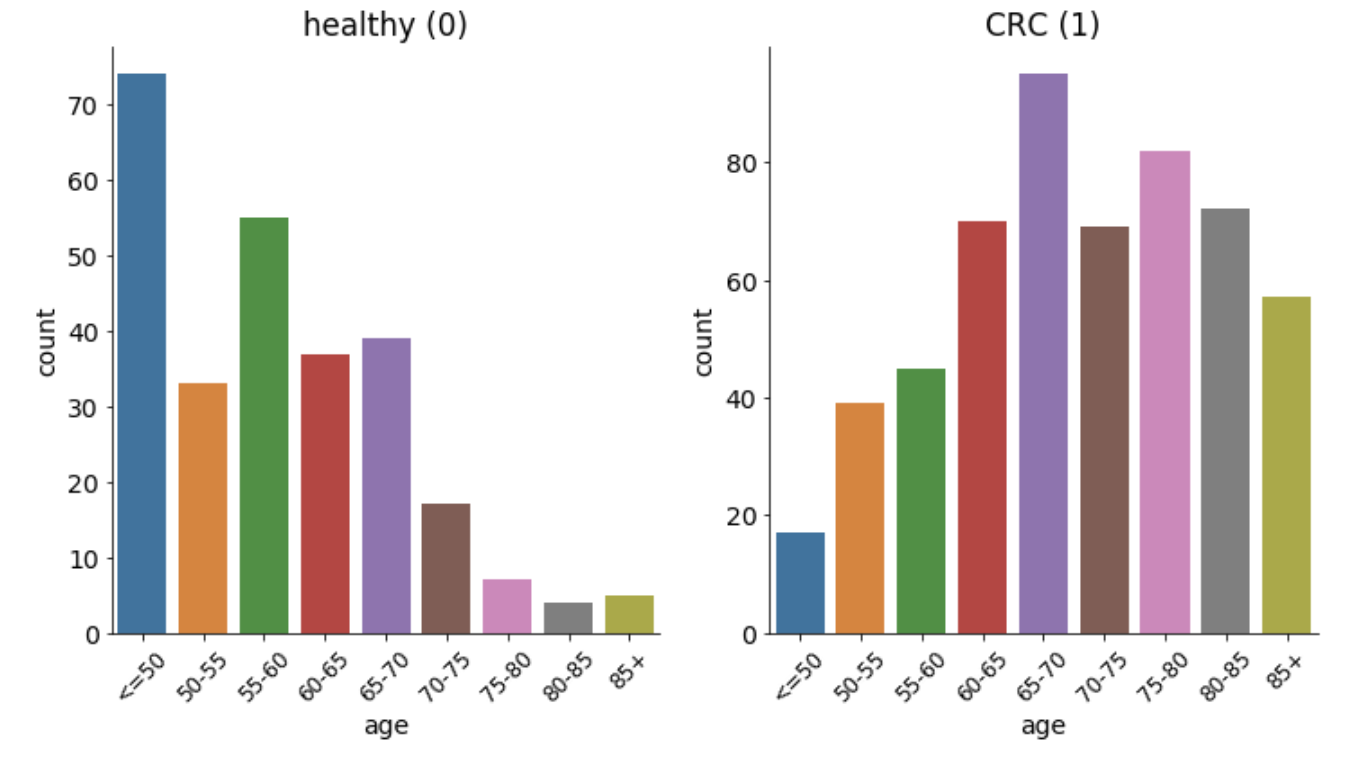}
    \caption{Age bin distribution split by class label}
    \label{fig:age_distribution}
\end{figure}

The distribution of samples is not uniform across the disease labels nor the confounders.  In this dataset, there are more colorectal samples than not.  There is a total of 9 institutions and 50 batches. Note the stark differences in the distribution of ages in Figure \ref{fig:age_distribution}.

\subsection{Hyperparameter selection} \label{hyperparameter_selection}

We selected hyperparameters for HCP and \Methodname\ by running a small grid search over a held-out data set.
For HCP we swept over the regularization parameters of $B$, the number of
components in the unknown covariate matrix, $X$, 
and the contribution of the $XW$ and $FB$
term to the loss function. For \Methodname\ all networks analyzed consisted of a hidden layer 
of ReLU neurons. The number of neurons and dropout probability were analyzed.

\subsection{Visualization of embeddings} \label{manifolds}

\begin{figure}
    \centering
    
    \begin{subfigure}[b]{0.4\textwidth}
    \includegraphics[width=\textwidth]{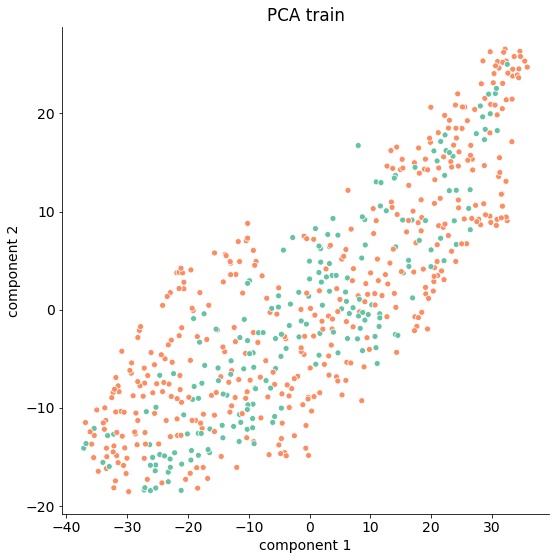}
    \label{fig:pca_label_train}
    \end{subfigure}
    \begin{subfigure}[b]{0.4\textwidth}
    \includegraphics[width=\textwidth]{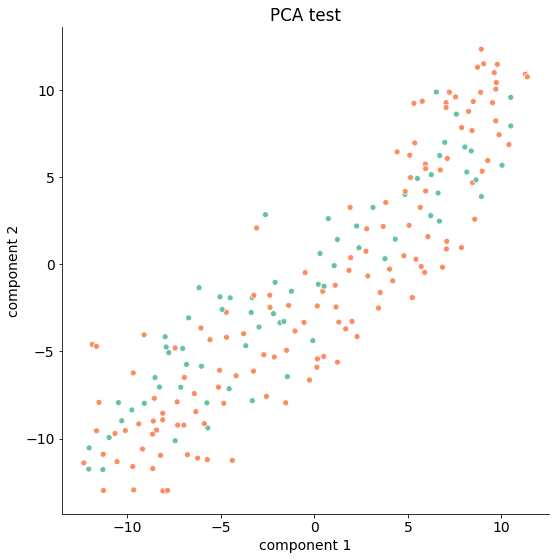}
    \label{fig:pca_label_test}
    \end{subfigure}
    
    \begin{subfigure}[b]{0.4\textwidth}
    \includegraphics[width=\textwidth]{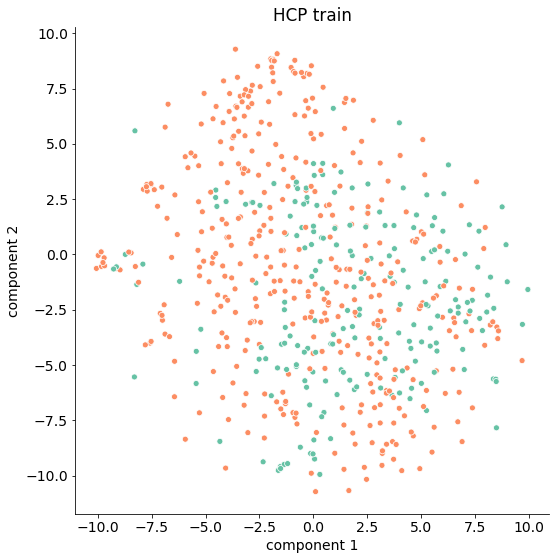}
    \label{fig:hcp_label_train}
    \end{subfigure}
    \begin{subfigure}[b]{0.4\textwidth}
    \includegraphics[width=\textwidth]{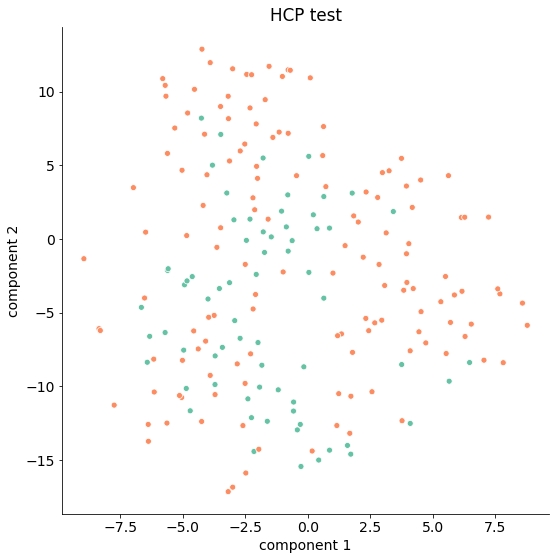}
    \label{fig:hcp_label_test}
    \end{subfigure}
    
    \begin{subfigure}[b]{0.4\textwidth}
    \includegraphics[width=\textwidth]{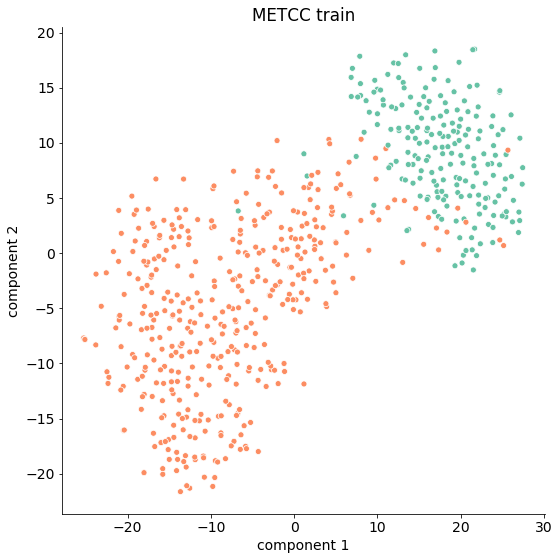}
    \label{fig:trip_label_train}
    \end{subfigure}
    \begin{subfigure}[b]{0.4\textwidth}
    \includegraphics[width=\textwidth]{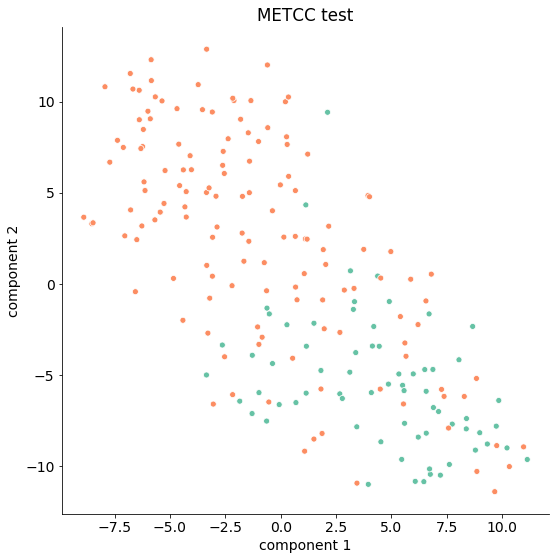}
    \label{fig:trip_label_test}
    \end{subfigure}

    \caption{Separate TSNE of train and test sets for one of the folds for each embedding with sample's label indicated by color.  Note the clear separation with METCC.}
    \label{fig:label_embedding}
\end{figure}

\begin{figure}
    \centering

    \begin{subfigure}[b]{0.4\textwidth}
    \includegraphics[width=\textwidth]{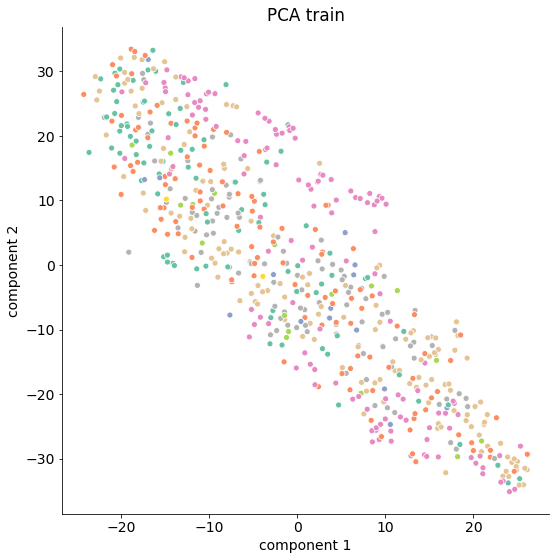}
    \label{fig:pca_institute_train}
    \end{subfigure}
    \begin{subfigure}[b]{0.4\textwidth}
    \includegraphics[width=\textwidth]{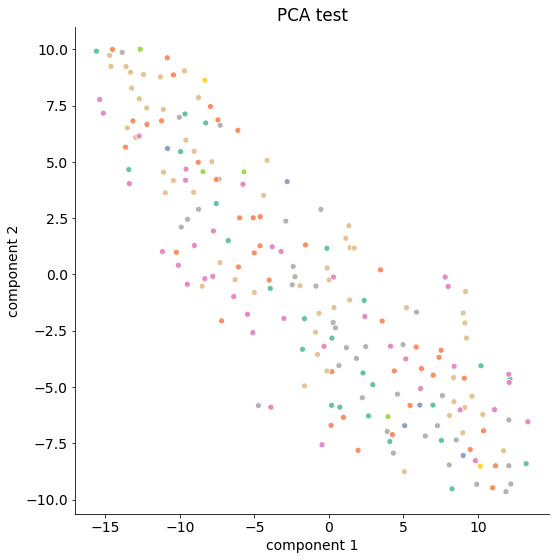}
    \label{fig:pca_institute_test}
    \end{subfigure}

    \begin{subfigure}[b]{0.4\textwidth}
    \includegraphics[width=\textwidth]{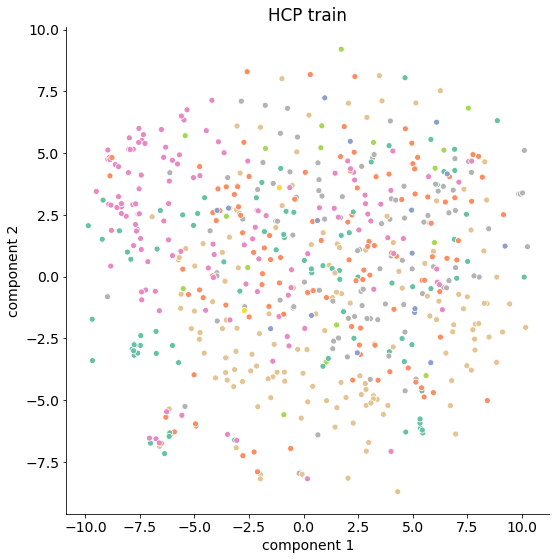}
    \label{fig:hcp_institute_train}
    \end{subfigure}
    \begin{subfigure}[b]{0.4\textwidth}
    \includegraphics[width=\textwidth]{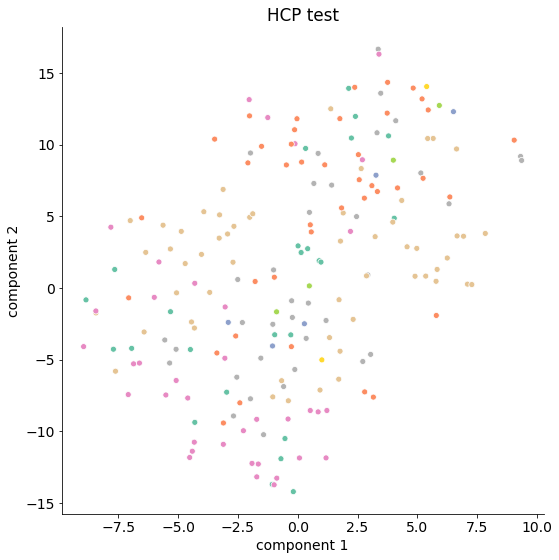}
    \label{fig:hcp_institute_test}
    \end{subfigure}

    \begin{subfigure}[b]{0.4\textwidth}
    \includegraphics[width=\textwidth]{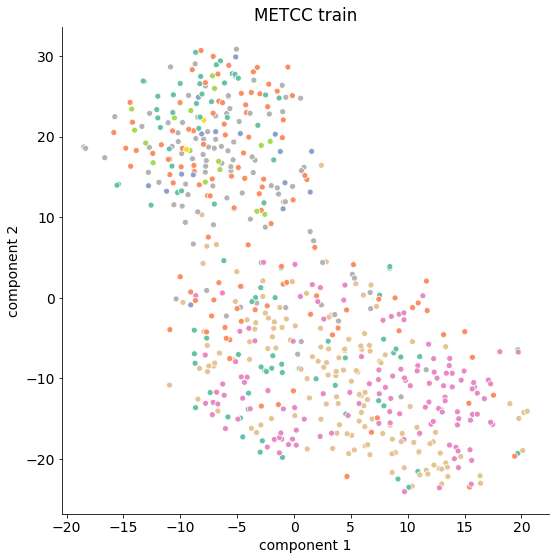}
    \label{fig:trip_institute_train}
    \end{subfigure}
    \begin{subfigure}[b]{0.4\textwidth}
    \includegraphics[width=\textwidth]{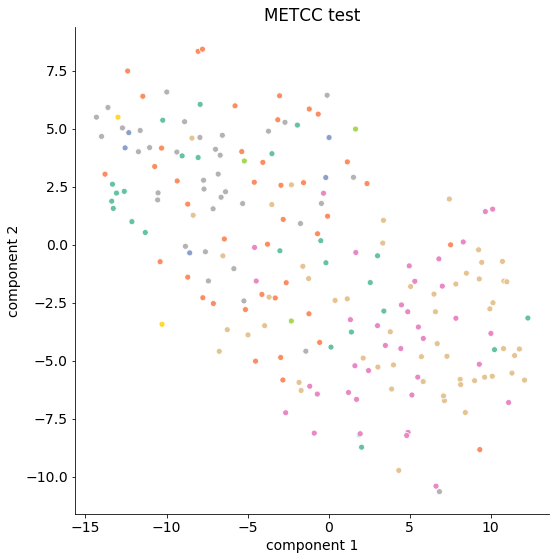}
    \label{fig:trip_institution_test}
    \end{subfigure}

    \caption{Separate TSNE of train and test sets for one of the folds for each embedding with sample's institution indicated by color.  Separability here could indicate confounding signal.}
    \label{fig:institution_embedding}
\end{figure}

\begin{figure}
    \centering
    
    \begin{subfigure}[b]{0.4\textwidth}
    \includegraphics[width=\textwidth]{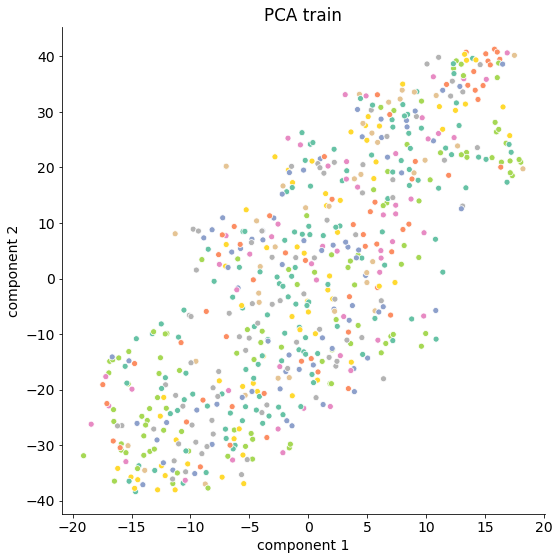}
    \label{fig:pca_flowcell_train}
    \end{subfigure}
    \begin{subfigure}[b]{0.4\textwidth}
    \includegraphics[width=\textwidth]{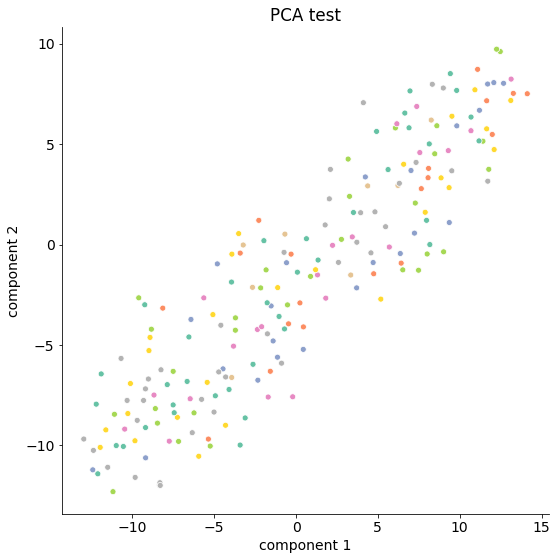}
    \label{fig:pca_flowcell_test}
    \end{subfigure}
    
    \begin{subfigure}[b]{0.4\textwidth}
    \includegraphics[width=\textwidth]{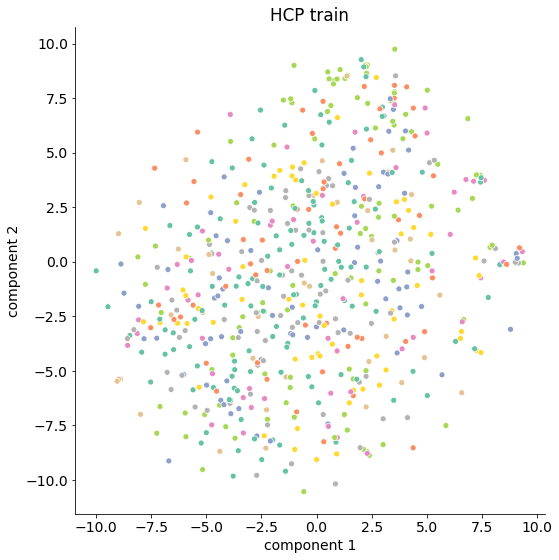}
    \label{fig:hcp_flowcell_train}
    \end{subfigure}
    \begin{subfigure}[b]{0.4\textwidth}
    \includegraphics[width=\textwidth]{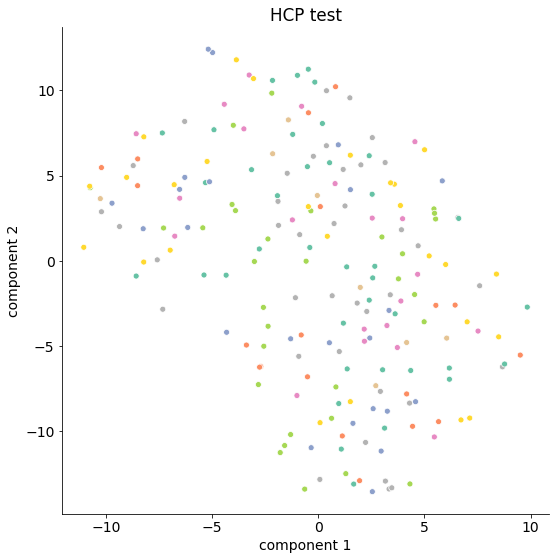}
    \label{fig:hcp_flowcell_test}
    \end{subfigure}
    
    \begin{subfigure}[b]{0.4\textwidth}
    \includegraphics[width=\textwidth]{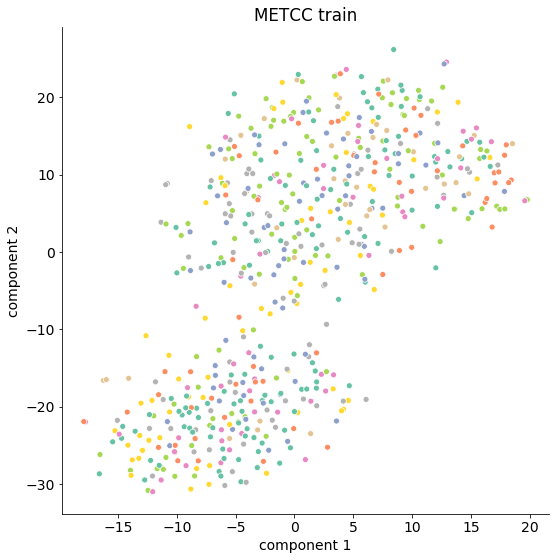}
    \label{fig:trip_flowcell_train}
    \end{subfigure}
    \begin{subfigure}[b]{0.4\textwidth}
    \includegraphics[width=\textwidth]{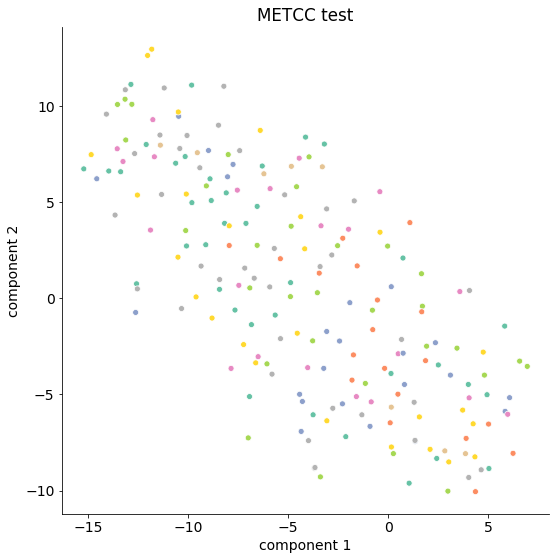}
    \label{fig:trip_flowcell_test}
    \end{subfigure}

    \caption{Separate TSNE of train and test sets for one of the folds for each embedding with sample's batch indicated by color.  Separability here could indicate confounding signal.}
    \label{fig:flowcell_embedding}
\end{figure}

\begin{figure}
    \centering
    
    \begin{subfigure}[b]{0.4\textwidth}
    \includegraphics[width=\textwidth]{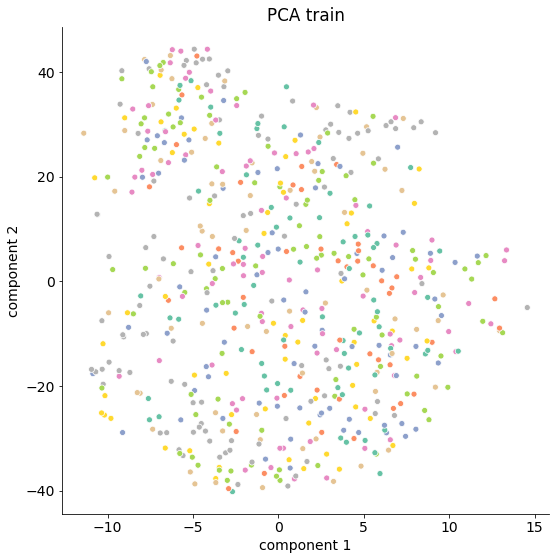}
    \label{fig:pca_age_train}
    \end{subfigure}
    \begin{subfigure}[b]{0.4\textwidth}
    \includegraphics[width=\textwidth]{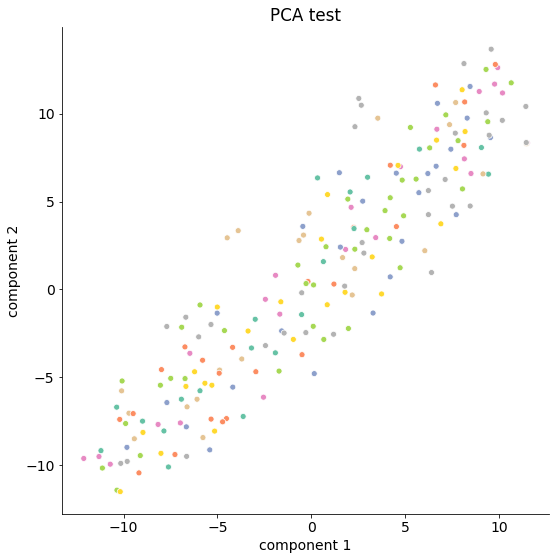}
    \label{fig:pca_age_test}
    \end{subfigure}
    
    \begin{subfigure}[b]{0.4\textwidth}
    \includegraphics[width=\textwidth]{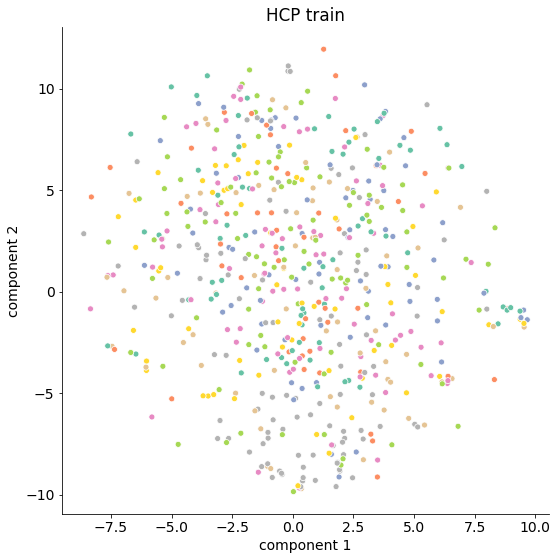}
    \label{fig:hcp_age_train}
    \end{subfigure}
    \begin{subfigure}[b]{0.4\textwidth}
    \includegraphics[width=\textwidth]{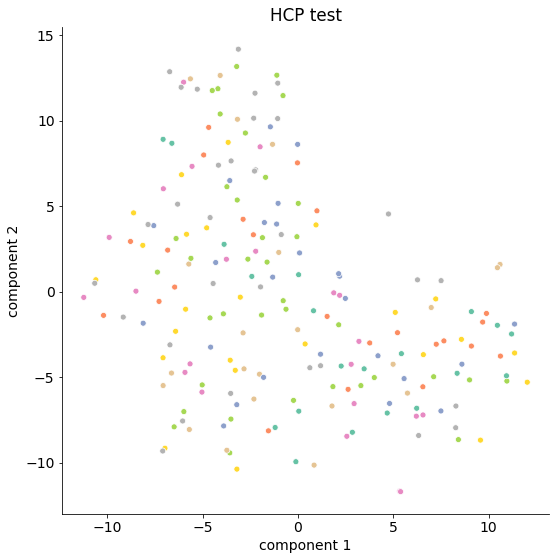}
    \label{fig:hcp_age_test}
    \end{subfigure}
    
    \begin{subfigure}[b]{0.4\textwidth}
    \includegraphics[width=\textwidth]{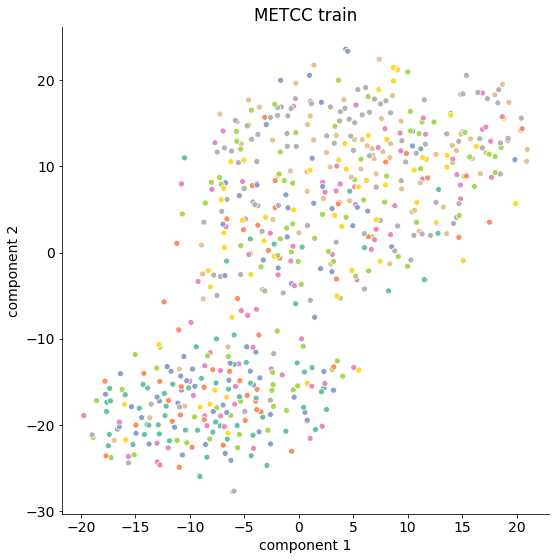}
    \label{fig:trip_age_train}
    \end{subfigure}
    \begin{subfigure}[b]{0.4\textwidth}
    \includegraphics[width=\textwidth]{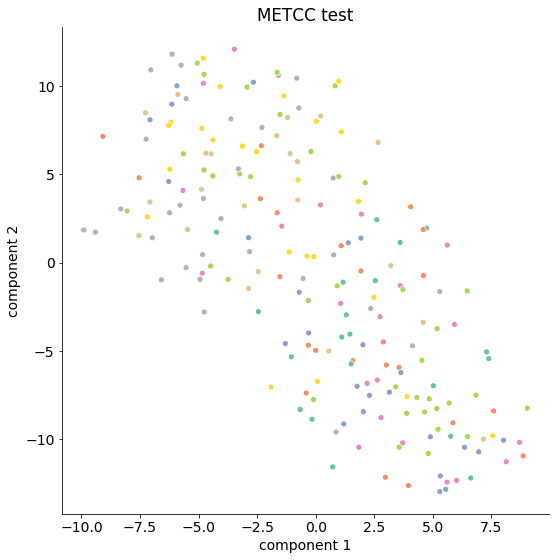}
    \label{fig:trip_age_test}
    \end{subfigure}

    \caption{Separate TSNE of train and test sets for one of the folds for each embedding with sample's age bin indicated by color.  Separability here could indicate confounding signal.}
    \label{fig:age_embedding}
\end{figure}
We applied TSNE to the embeddings $X_{PCA}$, $X_{HCP}$, $X_{\Methodname}$ \cite{scikit-learn}. We colored the projected data by the disease sample (Figure \ref{fig:label_embedding}), as well as by the confounders being assessed, institution (Figure \ref{fig:institution_embedding}), batch (Figure \ref{fig:flowcell_embedding}) and age (Figure \ref{fig:age_embedding}).


\end{document}